\documentclass[11pt,a4paper]{article}
\usepackage[hyperref]{emnlp-ijcnlp-2019}

\usepackage{amsmath}
\usepackage{amsfonts}
\usepackage{times}
\usepackage{latexsym}
\usepackage{multirow}
\usepackage{booktabs}
\DeclareMathOperator*{\argmax}{argmax}
\usepackage{url}
\usepackage{verbatim}

\title{Why can't memory networks read effectively?\Thanks{The code is available at {\footnotesize\url{http://bit.ly/2MOK6a4}}.}}
\author{Simon \v{S}uster \hspace{0.8cm} Madhumita Sushil \hspace{0.8cm} Walter Daelemans\\
  Computational Linguistics \& Psycholinguistics Research Center,\\ University of Antwerp, Belgium \\
  {\tt \{simon.suster,madhumita.sushil,walter.daelemans\}@uantwerpen.be}}
\date{}

\begin{document}

\maketitle
\begin{abstract}
   Memory networks have been a popular choice among neural architectures for machine reading comprehension and question answering. While recent work revealed that memory networks can't truly perform multi-hop reasoning, we show in the present paper that vanilla memory networks are ineffective even in single-hop reading comprehension. We analyze the reasons for this on two cloze-style datasets, one from the medical domain and another including children's fiction. We find that the output classification layer with entity-specific weights, and the aggregation of passage information with relatively flat attention distributions are the most important contributors to poor results. 
   We propose network adaptations that can serve as simple remedies.
   We also find that the presence of unseen answers at test time can dramatically affect the reported results, so we suggest controlling for this factor during evaluation.
   
   
\end{abstract}

\section{Introduction}



Recent work in machine reading comprehension and question answering has put a lot of focus on proposing new datasets and developing new reading architectures.
Among the latter, memory networks \citep{weston2014memory,sukhbaatar2015end,MillerEtAl2016} have played a prominent role, spanning the research on general reading comprehension, such as using children's fiction \citep{HillEtAl2016} and news texts \citep{KadlecEtAl2016}, as well as the research on knowledge integration and reasoning \citep{kumar2016ask,DasEtAl2017,kaushik2018much,chendurrett2019how}.\footnote{Many architectures, often with more complex memory-organization mechanisms, have been inspired by the original memory networks, e.g.\ \citet{graves2016hybrid}, \citet{weissenborn2016separating}, \citet{xiong2016dynamic}, \citet{hudson2018compositional}.}

While the workings of memory networks for reading comprehension and question answering have already been intensively studied, the emphasis has so far gravitated mainly around their ability to peform \textit{multi-hop reasoning}, where the network is expected to combine evidence from different pieces of text when deriving the correct answer. The main conclusion has been that memory networks have great difficulties learning multi-hop reasoning in an end-to-end way \citep{chendurrett2019how,chendurrett2019}. Chen and Durrett have demonstrated that even after providing a strong supervision signal to the attention component, the results on WikiHop \citep{welbl2018constructing} and bAbI \citep{weston2015towards} benchmarks still remained far behind other competing models.
Our work complements their work and extends it by focusing on the applicability and effectiveness of memory networks in \textit{single-hop} rather  than multi-hop machine reading comprehension. Additionally, whereas Chen and Durrett's work focused primarily on memory network's use of attention and on the proposed solution of attention supervision, we investigate a feature-based approach to give the network a strong attention signal, and also discuss other factors that lead to network's ineffectiveness.  


In the first part of the paper, we describe our attempt (\autoref{sec:clicr}) at establishing competitive results with memory networks on a medical reading comprehension task where other baselines and neural sequential models achieve much better results. Attracted by its transparent and extensible architecture, we had planned to extend memory networks for knowledge integration in the medical domain. Instead, we have found them to be ineffective at plain reading comprehension, so we have set out to explore the reasons behind this in more depth. We identify two major factors. First, the large parametrized output layer suffers from unseen answers at test time, which can be alleviated by either including a pointing layer, which obviates the need to keep weights for each answer entity, or performing entity anonymization, which reduces the size of the output layer and thus sparsity. Second, aggregating vectors of different passage windows into a single representation is harmful. Since the attention weights are distributed rather evenly, the most relevant passage windows for the given query aren't given enough prominence in the aggregated representation. 
We can reduce this problem by providing a stronger signal to the network about what passage window has the highest attention weight.


We also have a closer look (\autoref{cbt}) at the scenario where memory networks have been previously shown to work well, most clearly on Children's Book Test (CBT) \citep{HillEtAl2016}. We find that we can indeed obtain competitive results using the same architecture and implementation. However, through experimentation we confirm the findings from the literature that the scores on CBT are inflated due to a dataset design bias, and argue that memory networks don't effectively take into account the full evidence presented to them.

With our analysis, we are adding to the existing body of work that is (re-)examining the established machine reading and language understanding architectures, e.g.\ \citet{WangEtAl2017,kaushik2018much,chendurrett2019how}. We hope that our results can be of use also more broadly to researchers using other architectures or tasks.



\section{Models and datasets}
Our cloze-style reading comprehension tasks involve predicting a missing entity as the answer to a query $Q$, which includes a gap. A support text passage $P$ is given in which the answer can be found. 


\subsection{Memory networks}
We first concisely describe the generic architecture of the memory network (MemNet) \citep{weston2014memory,sukhbaatar2015end} used in our work. In a window-based representation, the passage $P$ consists of windows around candidate answer entities, and the query $Q$, which is also represented as a window around the empty slot. The words in the passage windows and the query are mapped into $d$-dimensional vectors using an embedding matrix $E\in\mathbb{R}^{|V|\times d}$, where $|V|$ is the size of the vocabulary. 
The query and passage window (memory slot) \textbf{encodings} are then obtained by averaging:
\begin{equation}
    q = \frac{1}{|Q|}\sum_{i\in Q} q_i;\hspace{0.5cm} p_j = \frac{1}{|P_j|}\sum_{k\in P_j} p_k,
\end{equation}
\noindent where $q_i$ is the embedding of the word at position $i$ in the query window, and $p_k$ is the embedding of $k$th word in $j$th passage window. The words in the passage window include the candidate answer entity.

Next, an attention mechanism measures the \textbf{compatibility} between the query representation $q$ and the window representations $p_j$. We formulate the attention as:
\begin{equation}
\alpha_j = softmax(cos(q,p_j)),
\end{equation}
\noindent where $cos$ is the cosine similarity.\footnote{We have investigated several other possibilities, including parametrized attention (cf.\ \citet{seo2016bidirectional}), and finally chosen the cosine similarity as the simplest yet a well performing function.} This gives us an attention weight for every slot $j$, indicative of its importance with respect to the query. The attention probabilities are then used to weigh the contribution of each memory slot into an \textbf{aggregate output} vector: 
\begin{equation}
o = \sum \alpha_j p'_j,
\end{equation}
\noindent where $p'$s are obtained using another embedding matrix $E'\in\mathbb{R}^{|V|\times d}$. This matrix is different from $E$ as we would like to distinguish between the representation used for measuring passage-query compatibility and the representation used for answering.
In the key-value variant \citep{MillerEtAl2016,kaushik2018much}, $p_j$ only represents a window encoding around the candidate,  and $p'_j$ an encoding of the candidate itself. Empirically, this doesn't lead to improvements in our case, so we don't use it. 

Finally, in a single-hop architecture like ours, we feed the output vector $o$, the query vector $q$, and their additive and multiplicative interactions directly to the \textbf{output classifier}, which is a softmaxed linear layer:
\begin{equation}\label{eq:output}
softmax(W([o; q; o+q; o\odot q]) + b),
\end{equation}
\noindent where $W\in\mathbb{R}^{C\times 4d}$, $C$ is the number of output labels, or answer candidates, and $b$ is the bias term. Although the original architecture feeds only $o+q$ to the output classifier, we have found our approach to be empirically superior. 

\subsection{Medical machine reading: CliCR}\label{sec:clicrdataset}
The dataset of English clinical case reports (CliCR) \citep{suster-daelemans-2018-clicr} is a gap-filling reading comprehension dataset consisting of around 100,000 queries and their associated documents. The dataset was built from case reports, requiring the machine reader to answer the query with an entity which is either a medical problem, a test or a treatment. Unlike in the original paper, we first pre-process the dataset by recreating the data splits so that those instances whose answer isn't found literally in the passage text are removed. This increases the scores across the board, and, crucially, ensures that pointing-based models can also be applied to the dataset. We rerun the baselines and comparison models from the original paper, and report those in Table~\ref{tab:clicrresults}. 

\paragraph{Baselines}
The baselines include choosing as answer a random entity in the test passage (\textbf{random}) and selecting the most frequent passage entity (\textbf{max-freq}). We also include a distance-based method that uses pre-trained word embeddings (\textbf{sim-entity}), where we vectorize the passage and the query, and then choose that entity from the passage whose representation has the highest cosine similarity to the query representation. These are the same baselines as reported in \citet{suster-daelemans-2018-clicr}. 

\paragraph{Other neural readers}
For comparison, we include several other neural systems. In the Stanford Attentive reader (SAReader) \citep{ChenEtAl2016} and the Gated-Attention reader (GAReader) \citep{DhingraEtAl2017a}, the architectures resemble that of MemNet, but the query and the passage are encoded using bidirectional GRUs. While SAReader predicts the answer with a classification layer, as in MemNet, GAReader drops it and bases the answer prediction directly on attention weights using a pointer mechanism \citep{Vinyals2015,KadlecEtAl2016,wang2016matchlstm}.

We additionally apply two models whose results haven't yet been published on CliCR. The first is BiDAF \citep{seo2016bidirectional}, which represents input words using a character-level CNN and pretrained word embeddings, and then encodes the query and passage with an LSTM. Both query-to-passage and passage-to-query attention is computed, but it is not used to summarize the passage into a fixed-size vector. The resulting vectors from the LSTM and the attention part are then passed to the recurrent modeling layer.
We also apply QANet \citep{yu2018qanet}, which encodes words similarly as BiDAF, but whose modeling components are based on the idea of stacked encoding blocks, which consist of CNN, self-attention and feed-forward layers. Both BiDAF and QANet predict an answer span using a pointing mechanism.

\subsection{Narrative machine reading: Children's Book Test (CBT)}\label{sec:cbtdataset}
To contrast the results obtained on CliCR with those from another gap-filling dataset, we use Children's Book Test (CBT) \citet{HillEtAl2016}, which consists of passages from children's literature in English. Each instance contains 21 consecutive sentences, where the last sentence represents a query with a missing slot. The goal is to fill the slot with either a named entity (NE), a common noun (CN), a verb (V) or a preposition (P), depending on the task. For each query, a list of 10 candidates is already given.
We apply the same baselines as with CliCR.

\section{Analysis of the failure case}\label{sec:clicr}
In this section, we analyze the performance of MemNet on CliCR.

We have trained the network for 10 epochs and selected the best performing model based on the development set accuracy. We have tuned the learning rate of Adam \citep{kingma2014adam} to $\{0.01,0.005,\underline{0.001},0.0005\}$, the embedding dimensionality to $\{50,\underline{100},200\}$, the number of hops to $\{\underline{1},2,3\}$, and the window size to 2 to each side (thus in total 5 tokens including the entity). We have set the size of the memory to 300. The word embeddings were pretrained on a combination of the training set of CliCR and PubMed abstracts \citep{hakala2016pubmed}, amounting in total to over 9 billion tokens. Other details about embedding pretraining can be found in \citet{suster-daelemans-2018-clicr}. 

We summarize the results in Table \ref{tab:clicrresults} and discuss them in turn in this and the following section.
The first thing to note is that although we can fit the training data well with MemNet, the generalization to unseen data remains poor as we only obtain 16.8 F1 on the test set. This is much lower than that for other neural readers, and even lower than that for the maximum frequency and embedding baselines. Simple modifications such as changing the definition of the memory from window-based to either sentences or key-value pairs have no positive effect.
We now present the factors that strongly affect this low performance.

\paragraph{Effect of unseen answers} \label{sec:pretrainedoutput}

We first have a look at the number of training instances available for answer labels. For those answers at test time that are also observed in the training set, the median frequency is 4 (mean=12, std.\ dev.=39, as measured on the development set). However, as much as around 60\% of all answers in the development set are never observed as answers in the training set.\footnote{All entities from the training set build up the possible output answer space. At test time, the correct answer may have occurred as an entity in a training passage, but not as an answer.} 
To investigate how this affects the results, we subsample a new test set (called \textit{seen}), which satisfies the condition that all data instances must contain answers which are already observed in the training set as entities. 
We see a considerable improvement on this reduced test set---compared to the original performance of 16.8 F1, we now achieve an F1 of 25.0. This indicates that the low performance can be partly attributed to the unseen answers in the test set. In general, we expect this effect to be amplified in situations exactly like ours, where the output layer is large (about 567 thousand entities). Interestingly, however, the unseen answers do not present a problem to other neural readers, which can be explained by a different way that predictions are achieved in these readers. More precisely, for GAReader, BiDAF and QANet, it is irrelevant at test time whether the true answer has been observed previously or not since the prediction is a \textbf{pointer} to the passage and not an answer entity from the fixed output vocabulary. In this way, the prediction is typically achieved based on attention directly (selecting as answer the entity whose window has the highest attention weight), without the need to learn the representation for each answer label in the output space. 
In the case of SAReader, although it still uses an output layer with dedicated weights for each answer, the widely adopted practice is to  \textbf{anonymize} the answer entities \citep{HermannEtAl2015,ChenEtAl2016,WangEtAl2017}. This procedure heavily reduces the output space and reduces the chance of observing a previously unseen answer at test time. It also forces the network to focus more on reading, since the anonymized entities (e.g.\ $@entity_3$) become effectively just pointers devoid of any semantics, apart from providing co-reference information \citep{WangEtAl2017}. As the results show, anonymization has an overwhelmingly positive effect for SAReader on CliCR.

To verify whether these alternative prediction mechanisms would have a positive effect on MemNet, we first try out a modification to its architecture where we replace the original output layer with a pointing layer, so that the prediction is a candidate from the window with the maximal attention weight. The F1 score in this case increases substantially, to 34.7. To apply anonymization, we keep the original output layer of MemNet but map all entities to numbered symbols, where the numbering is reinitiated for every data instance. For a given passage with $n$ different entity types, the set of entities is thus $\{\textit{@entity}_0, \textit{\ldots}, \textit{@entity}_{n-1}\}$. The final number of output labels in the network will correspond to the highest $n$ encountered in the training set.  In our case, this results in an output layer with a size of only 443 labels, where almost all are observed during training. With anonymization, the F1 score only increases marginally, from 16.8 to 18.1.

The above results suggest that the output layer as included in MemNet by default may not always be the best option, and that the presence of unseen answers at test time strongly influences the results.

\paragraph{The role of attention} Since the representations from MemNet's compatibility component are aggregated prior to being passed to the output layer, this could lead to blurring of information from several passage windows, especially if the attention weights aren't strongly peaked. In practice, we find that with a trained anonymized model, the average maximum attention probability is only 0.016 on the test set, with absolute variance from the mean at 0.009. This tells us that the attention distribution generally isn't strongly peaked towards any particular passage window. 

The results of MemNet-pointer and the embedding-similarity baseline suggest that contextual similarity between the query and a \textit{single} passage window should provide a relatively reliable clue about the correct answer, so it is reasonable to expect that a stronger signal from attention could help us to train a more competitive model. 
We therefore explore two options. The first uses a one-hot feature to indicate the best candidate (\textbf{attention-feat}):

\begin{align}
    \varphi&\in\{0,1\}^C: \sum_{i=1}^C \varphi_i=1,\\
    \varphi_i &=
    \begin{cases} 
   1 & \text{if } i=cand(\argmax_{j} \alpha_j), \\\nonumber
   0 & \text{otherwise.}
  \end{cases}
\end{align}
Here, the feature vector $\varphi_i$ is one exactly at the index corresponding  to the candidate answer whose window attained the maximum attention weight.\footnote{We also tried using soft probabilities as values of the feature vector, but this had a less positive effect.} 

This is a more direct way of biasing the model towards solutions preferred by hard selection than using attention supervision (cf.\ \citet{chendurrett2019how}). The feature vector is finally simply concatenated with other output vectors from eq.\ (\ref{eq:output}) to form the output representation. The output layer weights are expanded correspondingly, i.e.\ to $W\in\mathbb{R}^{C\times (4d+C)}$. Table~\ref{tab:clicrresults} shows that the attention feature significantly boosts the results, from 18.1 to 33.9 F1. When we use pretrained embeddings, this advantage is even greater at 36.7 F1. Clearly, the network can use the added feature to good effect, while aggregating without the added attention signal appears to be inadequate for good performance. An obvious next question is whether the aggregated information can contribute positively to the results \textit{at all}. If we drop the aggregated output vector and only keep the feature vector in the output layer (\textbf{attention-feat-only}), we observe a further improvement by two points, which could mean that the aggregated output vector is confusing the network, at least on these simpler instances which can be solved by using the attention feature alone. 

\begin{table}[t]
\centering
\small
\begin{tabular}{l l l}
Model & EM & F1 \\
\cmidrule(lr){1-3}

MemNet & 12.8 & 16.8 \\ 
\hspace{0.2cm}- \textit{seen} & 21.0 & 25.0 \\ 
\hspace{0.2cm}- ten-cands & 23.9 & 26.3 \\ 
\hspace{0.2cm}- ten-cands (\textit{seen}) & 55.7 & 57.5 \\
\hspace{0.2cm}- pointer & 27.5 & 34.7\\
\hspace{0.2cm}- anonymization & 13.3 & 18.1 \\ 
\hspace{0.4cm}- attention-feat & 27.8 & 33.9 \\ 
\hspace{0.4cm}- attention-feat (+emb-pretr) & 30.3 & 36.7 \\
\hspace{0.4cm}- attention-feat-only (+emb-pretr) & 32.2 & 38.7 \\
\hspace{0.4cm}- best-window & 30.3 & 36.8 \\
\hspace{0.4cm}- best-window (+emb-pretr) & 29.0 & 35.4 \\
QueryClassifier & 10.7 & 15.1\\
\hspace{0.2cm}- \textit{seen} & 16.8 & 21.4 \\
\cmidrule(lr){1-3}
random & 2.7 & 6.9 \\ 
max-freq & 14.6 & 19.0 \\ 
sim-window\textsuperscript{$\dag$} & 35.7 & 43.7 \\
Stanford Attentive Reader\textsuperscript{$\dag$} & & \\
\hspace{0.2cm}- anonymization & 34.0 & 41.3 \\
\hspace{0.2cm}- no-anonymization & 10.8 & 15.8   \\
Gated-attention reader\textsuperscript{$\dag$} & & \\
\hspace{0.2cm}- anonymization & 42.2 & 50.0 \\ 
\hspace{0.2cm}- no-anonymization & 38.2 & 45.6 \\
QANet & 39.5 & 46.5\\ 
BiDAF & 44.7 & 52.8 \\ 

\end{tabular}\caption{\textbf{CliCR} results. Results indicated with $\dag$ are for models from \citet{suster-daelemans-2018-clicr}. 
Results are reported for the test split with those questions not having an answer in the passage removed. The use of pretrained word embeddings for the model variants in the upper part of the table is indicated with \textit{+emb-pretr}, with randomly initialized word embeddings used otherwise. In the bottom part of the table, all neural models and \textit{sim-window} use pretrained embeddings.
Finally, for QANet, the data splits are smaller in size to prune out long sequences which led to memory issues.}
\label{tab:clicrresults}
\end{table}


Yet another possibility of adding a strong attention signal to the output layer, but without the use of the attention feature, is to avoid aggregation by feeding only the single-best passage window vector to the output layer (\textbf{best-window}):
\begin{equation}
    o = p'_i,\ \text{where}\ i = \argmax_j \alpha_j.
\end{equation}

\noindent This technique was also used by \citet{HillEtAl2016}, who call it \textit{self-supervision}.
We see here that we achieve results that are almost identical to those obtained with the attention feature. We can conclude that what appears to matter the most to the memory network is some indication of the best window, be it an indicator feature or the embedded window itself. A simple aggregate, however, is far inferior since the query-weighted passage representation gets too diffused as a consequence of flat attention weights. This probably means that the aggregate representation ends up resembling just an average over all passage windows. 

Overall, despite the fact that these modifications improve the network's performance considerably, the memory network remains a poor competitor to other machine reading systems in Table~\ref{tab:clicrresults}. The BiDAF reader, which adds additional complexity to the attention mechanism in comparison to SAReader and GAReader, performs the best on CliCR. \footnote{Although the readers that we compare to contain a larger number of parameters than MemNet, we don't observe reliable improvements when parametrizing MemNet more heavily (increasing the sizes of embedding and output matrices, and introducing a weight matrix into the attention function).}

\section{Contrasting with Children's Book Test}\label{cbt}
While our analysis from the previous section sheds some light on the workings of the memory network on CliCR,
we now include in our analysis also the CBT dataset, and compare some of its characteristics to CliCR. Strong performance has been previously shown on CBT using memory networks \citep{HillEtAl2016}. With our implementation that is identical to the one we have used on CliCR,\footnote{We have set the network's hyperparameters to the same values as on CliCR. The size of the memory was adjusted in such a way that all passage windows are used in training for each dataset part.} we obtain similarly strong performance as  \citet{HillEtAl2016} and \citet{kaushik2018much}. The results are included in Table~\ref{tab:cbtresults}.

We now discuss why the basic network without any modifications appears to work on CBT and not on CliCR. 

\paragraph{Effect of unseen answers} 
The first thing we note is that the performance on CBT varies a great deal across different parts of the dataset. The network does worst on the named entities part (NE).
The CBT dataset statistics in Table~\ref{tab:cbtstats} show that for named entities, only around \textit{a half} of the test answer types are observed in the training set, whereas for other parts this percentage is much higher.

\begin{table}[h]
\centering
\small
\begin{tabular}{l l l l l}
Model & NE & CN & V & P\\
\cmidrule(lr){1-5}
random & 10.0 & 10.0 & 10.0 & 10.0 \\
max-freq & 35.2 & 28.1 & 29.5 & 27.7 \\ %
sim-window & 31.4 & 26.0 & 30.7 & 23.2 \\
QueryClassifier & 43.1 & 45.3 & 67.2 & 63.5\\
\hspace{0.0cm}- \textit{seen} & 53.3 & 46.2 & 68.1 & 63.5\\ 
MemNet & 44.5 & 48.2 & 67.5 & 63.2 \\ 
\hspace{0.0cm}- \textit{seen} & 57.1 & 49.0 & 68.1 & 63.2 \\ 
\hspace{0.0cm}- \citet{HillEtAl2016} & 49.3 & 55.4 & 69.2 & 67.4 \\
\hspace{0.0cm}- \citet{kaushik2018much} & 35.0 & 37.6 & 52.5 & 55.2 \\
\end{tabular}\caption{Accuracy in \% on \textbf{CBT} for different dataset parts of the test set, using exact match. 
All MemNet models use random word embeddings.}\label{tab:cbtresults}
\end{table}

\begin{table}[h]
\centering
\small
\begin{tabular}{l l l l l}
 & NE & CN & V & P\\
\cmidrule(lr){1-5}
test size & 2500 & 2500 & 2500 & 2500  \\
ans types in train & 5078 & 4253 & 2322 & 85 \\
ans types in test & 422 & 696 & 406 & 43 \\
\% found in train & \textbf{56} & 93 & 93 & 100 \\
\end{tabular}\caption{CBT statistics, with a low proportion of named entities encountered as choices in the training set highlighted in bold.}\label{tab:cbtstats}
\end{table}

\paragraph{Number of candidates}
Furthermore, an obvious difference between the two datasets is that there are always ten candidates given in CBT, whereas in CliCR all entities in the passage are possible candidates (112 on average). To see the effect of reducing the pool of candidates on CliCR, we cap  the number of possible candidates to ten (as in CBT), thus including nine random candidates from the passage plus the correct answer.\footnote{It would also be possible to carry out an experiment where we adversarially add more false candidates to CBT and then stress-test on it.} This results in a dramatically higher score of 57.5 F1 compared to 25 F1 on the \textit{seen} test set of CliCR (Table~\ref{tab:clicrresults}). Of course, pre-selecting candidates makes the task less natural as we rely on an external oracle that shortlists the candidates. This practice, as performed on CBT, should generally be avoided as it is unreasonable to expect to have a strong test-time system for winnowing the candidates down without performing proper reading comprehension (cf.\ \citet{chendurrett2019} for a similar observation).

\paragraph{Query-only model}
Is the network exploiting a more direct correspondence between queries and answers in CBT? Prior work has shown that a baseline predictor which excludes passage information can return strong results on some datasets \citep{chendurrett2019how,chendurrett2019,Anand2018BlindfoldBF}. Our results confirm those of \citet{kaushik2018much}, where the query-only baseline reaches or surpasses the performance of the full MemNet.
This can explain why we see (seemingly) competitive results with MemNet. Despite a large proportion of unseen answers and the aggregation of window vectors, the network can still draw on the association between queries and answers, which constitutes a CBT's construction bias.


On CliCR, the original MemNet performance is already poor from the start, so we wouldn't expect to see a strong result when using the query input alone. Indeed, the query classifier obtains only 15.1 F1, with a gap to MemNet being more pronounced on the \textit{seen} test set (21.4 vs.\ 25.0 F1). In this respect, CliCR appears to be more difficult, as the network can't exploit the query-answer correspondences to obtain good results. Of course, the dataset could still contain other artefacts we are presently unaware of \citep{feng2019misleading}.



To return to \citet{kaushik2018much}'s study, their main finding in relation to the CBT dataset has been that it contains a bias where the query and, to a somewhat lesser extent, the passage can be used in isolation to predict answers with an unexpectedly high level of accuracy. 
We suspect that their results reveal not only the dataset bias, but also MemNet's poor reading capability, especially pertaining to measuring query-passage compatibility, and reworking this information successfully into an answer. Specifically, we see that other readers in \citet{kaushik2018much} obtain much higher results than the memory network, despite the bias. For example, GAReader records a forty-point advantage on the NE part, and 33 points on the CN part. This advantage might be explained by different text encoders of the two networks, which is a GRU in GAReader and an embedding layer in the memory network.
Also, the difference between the query-only model and the full model is 24.3 points (NE) for GAReader, but as little as 5.9 points for the memory network. We believe this points to memory network's modest capability of accounting for the passage in light of the query.


\section{Conclusion}
Our analysis reveals that using a vanilla variant of memory networks doesn't lead to competitive results on single-hop machine reading comprehension tasks.
We have identified two primary reasons for this. First, the use of the output classification layer that keeps weights for every answer candidate is detrimental. This appears to pose a problem in the presence of a large number of unseen answers at test time. We see that when we remove the instances with unseen answers from the test set, the results are improved dramatically. We have found that replacing the original output classification layer with a pointing mechanism or performing answer anonymization are better capable of dealing with the problem of unseen answers. Secondly, the aggregation of different window vectors with barely varying attention weights leads to a poor input representation for the final layer. Incorporating a stronger signal about the most compatible passage window, for example via a one-hot attention feature, leads to a perceivable improvement.





A few remarks about the limitations of our study are in place. The scope of our results covers a generic memory network only. For other memory network-based architectures with more complex encoding and compatibility layers the results might be different. While using pretrained word embeddings leads to an observable improvement, it is likely that other types of representations, such as pretrained contextualized embeddings from language models, may result in even stronger improvements. Finally, we have carried out the analysis on two cloze-style datasets only, which may limit the generalizability of our findings to other QA and machine reading datasets.

\section*{Acknowledgments}
This work was carried out in the framework of the Accumulate IWT SBO project (nr.\ 150056), funded by the government agency for Innovation by Science and Technology.
\bibliography{sample}
\bibliographystyle{acl_natbib}

\end{document}